\documentclass[letterpaper]{article}
\usepackage{natbib,alifeconf}
\usepackage{url}
\usepackage{framed,color}
\definecolor{shadecolor}{rgb}{.7, 1, .7}

\usepackage{placeins}

\title{Comparing Human and Automated Evaluation of Open-Ended Student Responses to Questions of Evolution}
\author{Michael J. Wiser$^{1,2}$, Louise S. Mead$^{1,2,3}$, James J. Smith$^{1,2, 3, 4, 5}$, \and Robert T. Pennock$^{1,2,4,6,7}$ \\
\mbox{}\\
$^1$BEACON Center for the Study of Evolution in Action  \\
$^2$Program in Ecology, Evolutionary Biology and Behavior, Michigan State University, East Lansing, MI, USA \\
$^3$Department of Integrative Biology, Michigan State University, East Lansing, MI, USA \\
$^4$Lyman Briggs College, Michigan State University, E. Lansing, MI, USA  \\
$^5$Department of Entomology, Michigan State University, East Lansing, MI, USA  \\
$^6$Department of Philosophy, Michigan State University, East Lansing, MI, USA  \\
$^7$Department of Computer Science \& Engineering, Michigan State University, East Lansing, MI, USA \\
mwiser@msu.edu}

\begin{document}
\setlength\titlebox{3.75in}
\maketitle

\begin{abstract}
Written responses can provide a wealth of data in understanding student reasoning on a topic.  Yet they are time- and labor-intensive to score, requiring many instructors to forego them except as limited parts of summative assessments at the end of a unit or course.  Recent developments in Machine Learning (ML) have produced computational methods of scoring written responses for the presence or absence of specific concepts.  Here, we compare the scores from one particular ML program -- EvoGrader -- to human scoring of responses to structurally- and content-similar questions that are distinct from the ones the program was trained on.  We find that there is substantial inter-rater reliability between the human and ML scoring.  However, sufficient systematic differences remain between the human and ML scoring that we advise only using the ML scoring for formative, rather than summative, assessment of student reasoning. 
\end{abstract}

\section{Background}


The central importance of evolution to teaching and learning in the biological sciences has been clearly established in all science education reform \citep{states_next_1900, brewer_vision_2011}.  Adequate formative assessment instruments -- administered during the course of instruction to gauge student understanding and reasoning in order to provide feedback for future instruction, instead of to assign a grade at the end of a unit -- that measure student understanding of evolutionary concepts \citep{bishop_student_1990, anderson_development_2002}, however, have until recently been rather limited \citep{nehm_measuring_2008}. Part of the challenge in designing an effective instrument comes from the fact that student understanding of evolutionary concepts is complex, and constantly changing. Studies find that students hold both scientifically accurate and naive or non-scientific explanations simultaneously \citep{andrews_biology_2012, hiatt_getting_2013} and that accurately identifying alternative conceptions can be difficult \citep{rector_learning_2012}.  Data also suggest students reason differently than experts, especially in response to different contextual elements of the sample questions.  Undergraduates employ more naive concepts when applying explanations of natural selection to plants as compared to animals; trait loss as compared to trait gain; and unfamiliar taxa as compared to familiar taxa \citep{nehm_item_2011}.  Furthermore, ascertaining the meaning of student responses is often very difficult.  One study found that 81 percent of students incorporated lexically ambiguous language in their responses to open ended questions about evolutionary mechanisms \citep{rector_learning_2012}.

Despite these challenges, assessing student knowledge is important, particularly in evaluating pedagogical practices designed to improve student understanding.   In an effort to identify effective assessment strategies, we have been investigating the applicability of a new tool, EvoGrader \citep{moharreri_evograder:_2014}.   
   
%
%
%

Open-ended student responses can provide a wealth of data about student reasoning.  Unfortunately, they can also be time- and labor-intensive to score.  One study found that it took an average of four minutes for a human grader to score a single response for the nine ideas we analyze in this study \citep{moharreri_evograder:_2014}.  For even a class of 30 students, scoring five such questions would take ten hours, which quickly becomes prohibitive.  If an instructor wants to get a general sense of student understanding on a formative assessment, a more rapid method is highly desirable.

	An appealing potential solution to this problem would be if instructors had an automated system that was sufficiently sophisticated to evaluate student answers to such open-ended questions.  Of course, this is not a simple task.  Even setting aside the difficulty of parsing open-ended natural language responses in general, one still has the further problem of interpreting the appropriateness of answers in relation to content knowledge and overarching concepts.  For instance, a science teacher may want to know whether a student's response demonstrates incorrect naive notions or whether it demonstrates concrete scientific understanding. Machine Learning systems have begun taking the first steps to accomplishing this difficult task.
    
\subsection{Use of Machine Learning in Education}

There is growing interest in using tools and techniques from Machine Learning in the classroom environment \citep{butler_integrating_2014}.  In fact, an entire book has been written about using Machine Learning in educational science \citep{kidzinski_tutorial_2016}.  One area of particular interest is language processing.  Machine learning techniques have been used to classify instructor questions according to Bloom's taxonomy \citep{yahya_analyzing_2013}.  Perhaps the biggest use of Machine Learning in an educational environment is in the automated scoring of student writing (reviewed in \citep{nehm_transforming_2012}). 

One domain-specific example of ML techniques in language processing is provided by the web portal EvoGrader, discussed below.  EvoGrader was designed to assess student understanding of natural selection, using a particular set of questions, consisting of a brief scenario and asking the students how a biologist would explain this scenario of evolutionary change or patterns. Our study seeks to measure how similar of scores this ML procedure provides to human scoring for  questions on which the application has not been trained but which are written in the same style.  

\subsection{EvoGrader}

EvoGrader (\url{http://www.evograder.com}) is a free, online service that analyzes open-ended responses to questions about evolution and natural selection, and provides users with formative assessments.  It is described in detail in \citep{moharreri_evograder:_2014}, but a brief description follows.

	EvoGrader works by supervised machine learning.  Participants (n=2,978) wrote responses to ACORNS assessment items \citep{nehm_reasoning_2012} and ACORNS-like items \citep{bishop_student_1990}, generating 10,270 student responses.  These items consist of a prompt describing a short scenario relevant to natural selection, and ask students to write how a biologist would explain this situation. Participants spanned many different levels of expertise, including non-majors, undergraduate biology or anthropology majors, graduate students, postdocs, and faculty in evolutionary science.  Each response was scored independently by two human raters for each of six Key Concepts (KC) and three Naive  Ideas (NI) (see Box 1).  These consensus scores were used to train EvoGrader, based on the supervised machine learning tools of LightSIDE \citep{mayfield_open_2013}.  LightSIDE provides feature extraction, model construction, and model validation, based on the human-scored responses.

	EvoGrader's authors chose different methods to optimize the scoring algorithm for feature extraction for the 9 scoring models (one model for each concept) -- all considered the dictionary of words used in a particular response, and reduced words to their stems; most removed high frequency low information words (e.g., the, of, and, it); some also included pairs of consecutive words (e.g., "had to", "passing on"), or removing misclassified data (see Moharreri et. al. \citep{moharreri_evograder:_2014} Table 2 for details). 

	After feature extraction, each response was converted to a set of vectors containing frequencies of words or word pairs.  These vectors were then passed to a binary classifier, which underwent Sequential Minimal Optimization (SMO) (Platt 1999) for each of the 9 models.  The SMO training algorithm iteratively assigned weights to words in the written responses until the model was able to match the human scores within a certain margin of error.  The models were then validated with 10-fold cross-validation, using 90\% of the data to generate a model and the remaining 10\% of the data to validate it, and then repeating this procedure for a total of 10 times such that each 10\% of the data was used for validation exactly once and model generation 9 times.  The authors averaged these models to get the final models used by the program, assessing whether they met quality benchmarks (90\% accuracy and kappa coefficients $\geq$ 0.8) defined by the creators, and adjusting the training until the models did.
%
%
%

\begin{figure}[htb]
	\centering
	\includegraphics[width=.9\columnwidth]{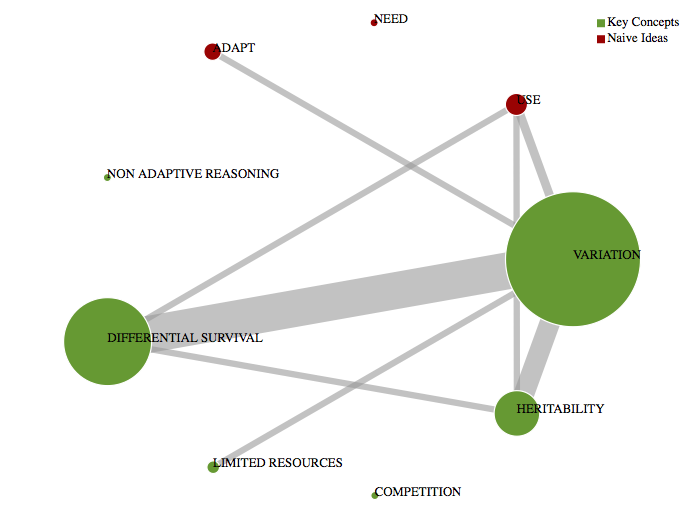}
    \includegraphics[width=.9\columnwidth]{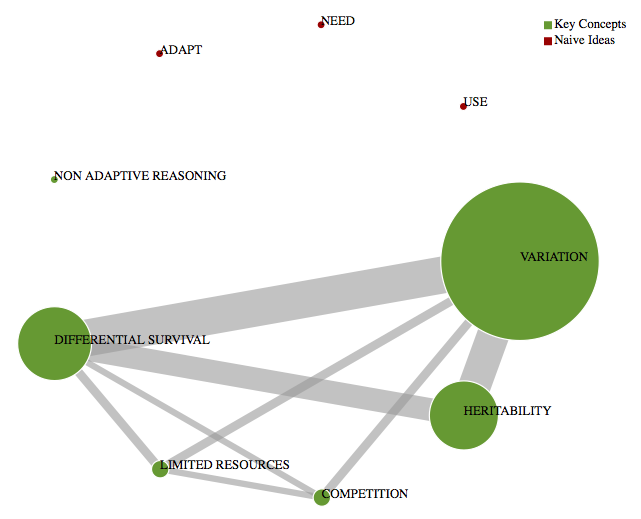}
    \caption{Concept maps produced by EvoGrader for the pre-instruction (upper panel) and post-instruction (lower panel) analysis of Question 1 (see Box 1).  Sizes of the circles indicate percentage of responses scored as containing that concept; widths of the lines connecting concepts shows frequency of co-occurrence of those concepts.}
    \label{fig:ABRPrePostEvoGrader}
\end{figure}

	EvoGrader uses these validated models to score new responses from web users.  Users must upload data in a specific format, which the portal verifies.  If the data is formatted correctly, EvoGrader then evaluates each response using the existing validated models, and provides both machine scored data in a downloadable .csv format and a variety of web visualizations of the data.  (Fig.  ~\ref{fig:ABRPrePostEvoGrader})
%
%
%
%
%

%
%
%
%
%
%
%
\section{Methods}
\subsection{Student data}
We administered pre-instruction and post-instruction tests consisting of two questions (see Box 1) about evolution to students in an Introductory Cell and Molecular Biology course in the fall semester of 2014. Both questions asked students about how evolutionary processes occur.  Question 1 asks about an evolutionary gain of antibiotic resistance in a population, while question 2 asks about the evolutionary loss of toxicity in a mushroom population.  Completed pre- and post-test responses were obtained from 34 students for question 1 and from 36 students for question 2.

\begin{shaded}
\section{Box 1}
We evaluated student responses to two prompts:\\

\noindent \textit{Question 1: Explain how a microbial population evolves resistance to the effects of an antibiotic.}\\

\noindent \textit{Question 2: A species of mushroom contains a chemical that is toxic to mammals. How would biologists explain the initial occurrence and increase in frequency of a number of individuals in the population that no longer produce this toxin?}\\

We scored each response for whether it contained each of the following concepts:\\

\textbf{Key Concepts}:
\begin{itemize}
{\item \textit{Variation}: The presence and causes (mutation/recombination/sex) of differences among individuals in a population.}
{\item \textit{Heritability}: Traits that have a genetic basis and are able to be passed on from parent to offspring.}
{\item \textit{Competition}: A situation in which two or more individuals struggle to get resources which are not available to everyone.}
{\item \textit{Limited Resources}: Required resources for survival (food, mates, water, etc) which are not available in unlimited amounts.}
{\item \textit{Differential Survival}: Differential survival and/or reproduction of individuals.}
{\item \textit{Non-adaptive Ideas}: Genetic drift and related non-adaptive factors contributing to evolutionary change.}
%
%
%
%
%
\end{itemize}

\textbf{Naive Ideas:}

\begin{itemize}
{\item \textit{Adapt}: Organisms/populations adjust or acclimate to their environment.}
{\item \textit{Need}: Organisms gain traits or advantage in response to a need or a goal to accomplish something.}
{\item \textit{Use/Disuse}: Traits are lost or gained due to use or disuse of traits.}
\end{itemize}

Further, human evaluators determined whether or not a response answered the question asked; if the response did not, no credit was given for Key Concepts.  For example, consider this student response: 
\begin{quotation}
\textit{“Similar to above, some kind of mutation for the poison and those plants were not eaten so they were able to reproduce and pass thoses} [sic] \textit{genes on to future generations. The population of poisonous mushrooms would soon outnumber non-poisonous ones since poisonous mushrooms are less likely to be eaten. Over time, animals would learn to stay away from teh} [sic] \textit{mushroom simply be [sic] appearance, so the toxin would no longer be needed.“}
\end{quotation}

Although this answer demonstrates adaptive reasoning about the origin of toxic mushrooms,  the question was about the loss of toxin in this population, not the origin of the toxin.  Only the last sentence addresses the loss of the toxin, and it does not demonstrate any of the Key Concepts.

\end{shaded}

\subsection{Data}
Data files containing all student responses, scoring, and data analysis may be found at \url{https://github.com/mjwiser/ALife2016}

\subsection{Scoring responses}
We used EvoGrader to score student responses on two open-ended questions about natural selection for six Key Concepts and three Naive Ideas (see Box 1).  Two human graders (MJW and LSM) scored student responses for these same criteria.  We resolved any disagreement among the humans by discussion, resulting in a consensus human score.

\subsection{Statistical analysis}
We measured inter-rater reliability (IRR) between the EvoGrader scores and the consensus human scores for each question, as outlined in \citep{hallgren_computing_2012}.   Because we were interested in the IRR of specific questions, we combined both pre-and post-instruction responses into a combined data set. We computed IRR both for each question as a whole, and separately for the key concepts and the naive ideas within each question.  We chose to not compute IRR for each individual concept, or separately for pre- and post-instruction questions, because of the lower statistical power from examining each set separately, and the increase in multiple comparisons this would necessitate.  We also compared the EvoGrader and human consensus scores by way of 2-tailed paired t-tests to test for differences in the number of key concepts or naive ideas scored.  We conducted all statistical testing in R version 3.2.3 \citep{r_core_team_r:_2013}.

\section{Results and Discussion}

\begin{figure*}[htb]
	\centering
	\includegraphics[trim = 40mm 0mm 80mm 0mm, clip, height=5in]{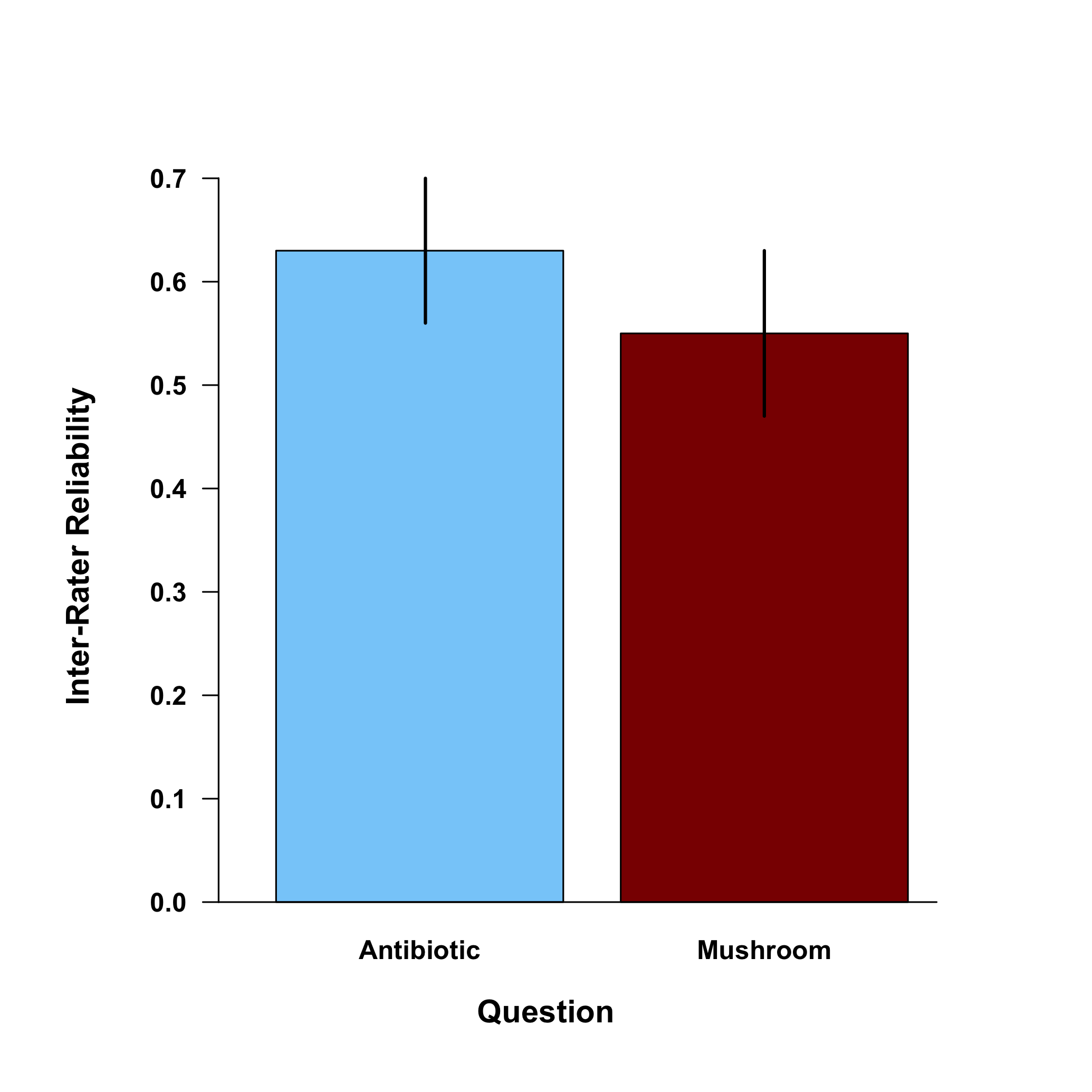}
    \caption{Inter-rater Reliability for Questions 1 and 2.  Key Concepts and Naive Ideas are pooled within each question.  Plotted values are Cohen's kappa.  Error bars shown are 95\% confidence intervals.}
    \label{fig:IRRbyquestion}
\end{figure*}
\begin{figure*}[tbp]
	\centering
	\includegraphics[trim = 40mm 40mm 100mm 0mm, clip, height=3in]{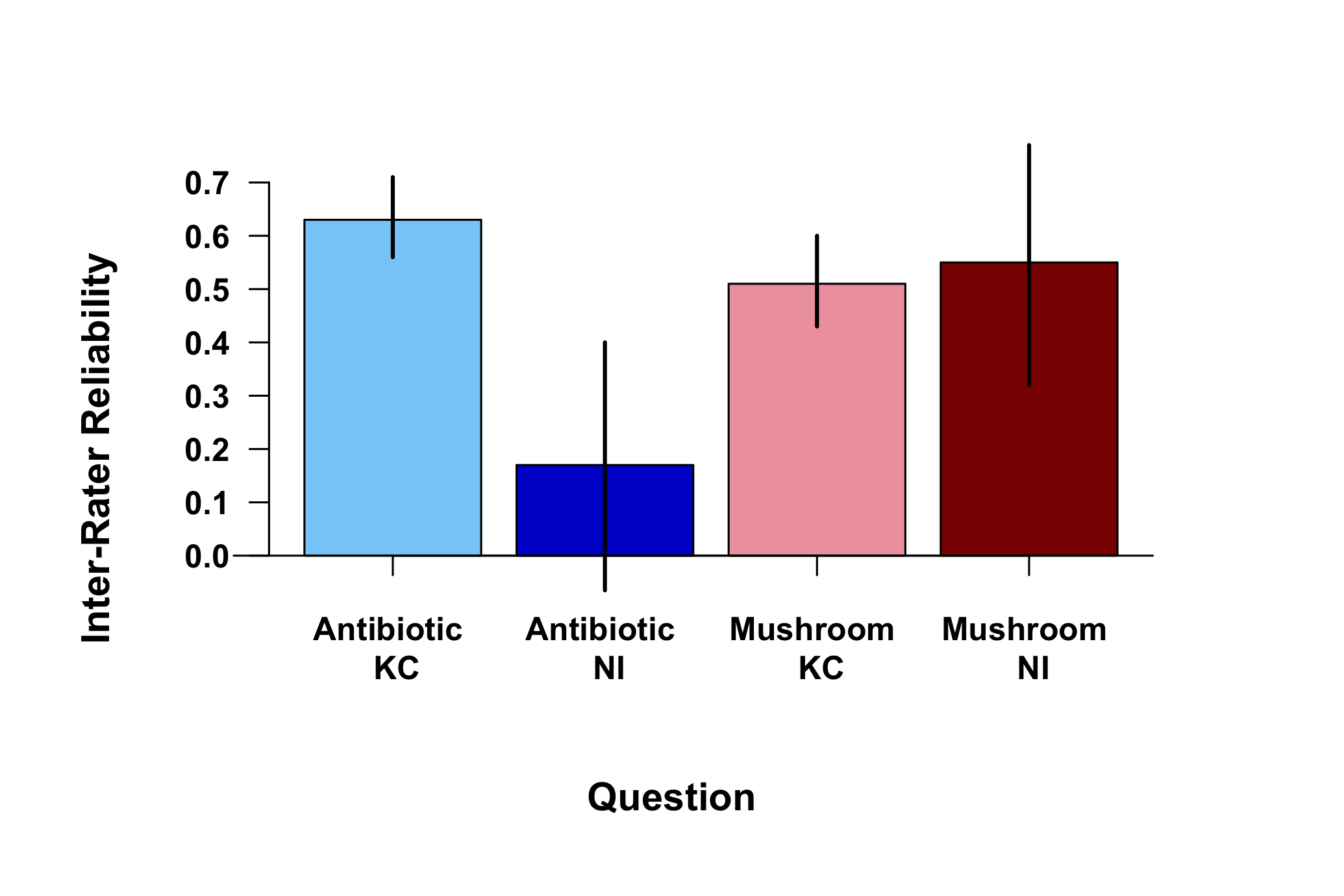}
    \caption{Inter-rater Reliability for Questions 1 and 2, broken down between Key Concepts (KC) and Naive Ideas (NI).  Plotted values are Cohen's kappa.  Error bars shown are 95\% confidence intervals.}
    \label{fig:IRRbytype}
\end{figure*}
	The Inter-Rater Reliability (IRR) of EvoGrader and the consensus human scoring of these questions is good, with values of 0.63 for the antibiotic resistance question and 0.55 for the mushroom question (Fig. ~\ref{fig:IRRbyquestion}).  This means that more than half of the total variance in scoring across these 9 concepts is shared among the raters.  Landis and Koch (1977) suggest that IRR values from Cohen’s kappa in the range of 0.6 to 0.8 indicate substantial agreement among coders, and values between 0.4 and 0.6 indicate moderate agreement \citep{landis_measurement_1977}.  By these criteria, when all of the concepts are analyzed together, the IRR for the antibiotic question is strong, and the IRR for the mushroom question is moderate.  
 	
   We further examined IRR separately for Key Concepts and Naive Ideas (Fig.  ~\ref{fig:IRRbytype}), to examine whether there was a systematic difference between the two concept types.  In the antibiotic resistance question, the IRR is notably higher for the Key Concepts than the Naive Ideas (0.63 v 0.17).  In fact, the 95\% confidence interval for the Naive Ideas IRR overlaps 0, meaning that the IRR is not statistically significantly different from ratings being assigned at random.  Conversely, IRR in the mushroom question is consistent across the Key Concepts and Naive Ideas (0.51 and 0.55, respectively), showing no meaningful difference across concept type.
%
%
%

	What can account for these differences in IRR?  One thing to take note of is that when there is very low variation in a given rater’s scoring across responses, there is very little statistical power to detect shared variance across raters.  As a thought experiment, imagine that two different raters assign scores of “Yes” to 10\% of responses, and “No” to 90\%.  Even if the two raters both assigned their scores randomly, the two raters would be expected to agree 82\% of the time.  IRR analyses take into account the expected frequency of scoring agreement, but a low variance across responses for a given rater will negatively affect the statistical power of IRR analyses.  This is reflected in the wide confidence intervals for the Naive Ideas in particular.  For one, there are fewer potential Naive Ideas scored (since there are at most three Naive Ideas per student response, while at most six Key Concepts per student response).  This skew in responses had a larger impact on the Naive Ideas in the antibiotic resistance question than elsewhere; EvoGrader only scored the entire class as expressing five total Naive Ideas in the antibiotic question; the consensus human score was 90.  This is part of a general trend: for both questions, the human consensus score differed from the EvoGrader score, and by a statistically significant margin even when correcting for multiple comparisons (see Table ~\ref{table:IRR_table}; all adjusted p-values \textless 0.05).  For both questions, the human consensus score detected more Naive Ideas than EvoGrader did.  However, the humans detected more Key Concepts than EvoGrader did for the antibiotic question (question 1), but fewer in the mushroom question (question 2).

\begin{table*}[tb]
\centering
\begin{tabular}{| c  r  c  c  c |}
  \hline					
	Comparison	& t  &  df & p & adj. p\\[2ex]  
  \hline
	Antibiotic KC & 5.779 & 67 & $2.14 * 10^{-7}$ & $8.58 * 10^{-7}$\\
  \hline
  	Antibiotic NI & 2.604 & 67 & $0.0113$ & $0.0453$\\
  \hline
  	Mushroom KC & -2.806 & 71 & $0.00647$ & $0.0259$\\
  \hline
  	Mushroom NI & 3.384 & 71 & $0.00117$ & $0.00466$\\
  \hline
\end{tabular}
\caption{2-tailed paired t-tests comparing EvoGrader and human consensus
scoring of Key Concepts (KC) and Naive Ideas (NI).  Negative values indicate
more of these concepts detected by EvoGrader; positive values indicate more
of these concepts detected by humans.  A Bonferroni correction was used to
generate the adj. p values. }
\label{table:IRR_table}
\end{table*}
	Several factors may serve to lower the IRR from ideal levels.  One obvious cause is mentioned in Box 1: some student responses demonstrate reasoning about natural selection, but do not answer the question asked.  In these cases, the humans did not credit the student with any of the Key Concepts that did not address the question asked.  EvoGrader, on the other hand, did not have this screening mechanism.  Further, we analyzed both pre- and post-instruction responses jointly, and we expect the number of Naive Ideas expressed to decrease through instruction while we expect the number of Key Concepts expressed to increase through instruction. 
Such instructional effects would be a positive outcome for students, but both may reduce variance in the post-instructional scoring, reducing the statistical power to detect shared variance.

 	What can account for the difference in results between the two questions?  There are two potentially salient contextual differences between the questions.  One, the first question is a gain of a trait, while the second is a loss of a trait.  Two, the two questions use different taxonomic groups as their examples.  Both of these differences have been shown in the literature to be important to student reasoning \citep{nehm_item_2011}.  In a future study, we will be able to disentangle these factors through a multifactorial design that considers multiple taxonomic groups and asks both a gain of trait and a loss of trait question within each. 

%
%
%

\section{Conclusions}
EvoGrader is a useful tool for assessing student reasoning about natural selection.  Even on questions not included in the training, it provides a reasonable level of reliability in scoring student responses on open-ended questions of a similar style to the ACORNS assessment.  However, it is not foolproof.  In our study, EvoGrader credited students as displaying more Key Concepts, and fewer Naive Ideas, than our human raters did. In particular, EvoGrader may inaccurately credit student responses that do not address the specific question asked for evolutionary reasoning.  For formative assessments, it can be a valuable tool to get a sense of student responses in a short period of time, but we caution against using EvoGrader to assign points to students, given its current limitations.
%
%
%

\subsubsection{Acknowledgments.} 
We thank Rohan Maddamsetti, Emily Dolson, Alex Lalejini, Anya Vostinar, Joshua Nahum, Brian Goldman, and Charles Ofria for helpful discussion during manuscript preparation.  This work was supported by the National Science Foundation IUSE No. 1432563 and under Cooperative Agreement No. DBI-0939454. Any opinions, findings, and conclusions or recommendations expressed in this material are those of the author(s) and do not necessarily reflect the views of the National Science Foundation.
\FloatBarrier

\bibliographystyle{apalike}
\bibliography{main}

\end{document}